\documentclass[conference]{IEEEtran}
\IEEEoverridecommandlockouts
\usepackage{cite}
\usepackage{amsmath,amssymb,amsfonts}
\usepackage{algorithmic}
\usepackage{graphicx}
\usepackage{textcomp}
\usepackage{xcolor}
\def\BibTeX{{\rm B\kern-.05em{\sc i\kern-.025em b}\kern-.08em
    T\kern-.1667em\lower.7ex\hbox{E}\kern-.125emX}}
\begin{document}

\title{Local Rose Breeds Detection System Using Transfer Learning Techniques\\
{\footnotesize \textsuperscript{}}

}
\author{\IEEEauthorblockN{Amena Begum Farha}
\IEEEauthorblockA{\textit{Department of CSE} \\
\textit{Daffodil International University}\\
Dhaka, Bangladesh \\
amena15-11845@diu.edu.bd}
\and
\IEEEauthorblockN{Md. Azizul Hakim}
\IEEEauthorblockA{\textit{Department of CSE} \\
\textit{University of Nevada}\\
Reno, USA \\
azizul.hakim@nevada.unr.edu}
\and
\IEEEauthorblockN{Mst. Eshita Khatun}
\IEEEauthorblockA{\textit{Department of CSE} \\
\textit{Daffodil International University}\\
Dhaka, Bangladesh \\
eshita.cse@diu.edu.bd}
\and
}

\maketitle


\begin{abstract}
Flower breed detection and giving details of that breed with the suggestion of cultivation processes and the way of taking care is important for flower cultivation, breed invention, and the flower business. Among all the local flowers in Bangladesh, the rose is one of the most popular and demanded flowers. Roses are the most desirable flower not only in Bangladesh but also throughout the world. Roses can be used for many other purposes apart from decoration. As roses have a great demand in the flower business so rose breed detection will be very essential. However, there is no remarkable work for breed detection of a particular flower unlike the classification of different flowers. In this research, we have proposed a model to detect rose breeds from images using transfer learning techniques. For such work in flowers, resources are not enough in image processing and classification, so we needed a large dataset of the massive number of images to train our model. we have used 1939 raw images of five different breeds and we have generated 9306 images for the training dataset and 388 images for the testing dataset to validate the model using augmentation. We have applied four transfer learning models in this research, which are Inception V3, ResNet50, Xception, and VGG16. Among these four models, VGG16 achieved the highest accuracy of 99\%, which is an excellent outcome. Breed detection of a rose by using transfer learning methods is the first work on breed detection of a particular flower that is publicly available according to the study.
\end{abstract}

\section{Introduction}
Bangladesh is known as the land of flowers. At recent, the flower cultivation covers 10,000 hectares of land. The number of farmers, who are growing flowers is more than 5,000 and around 150,000 people are engaged in flower business for their subsistence \cite{b1}. From the ancient time, people are using roses for many purposes.  Roses are the most desirable flowers for any kind of occasion. Rose water is the a very valuable thing that people are using according to their needs. Many perfumes have been made of rose oil. Besides making sweets, rose petals are being used in potpourri. Rose-hip is a fruit, which is abundant with vitamin c. In Bangladesh flower business in shortened without roses. The diversity of roses is improving day by day. Rosaceae is a plant family enlarged with over hundred genera and more thousands of species. As roses play a vital in flower business so knowing about all breeds of roses is really important. In Bangladesh, nine types of rose breeds have seen so far. These nine breeds of roses are Papa Meiland, Iceberg, Rose Gaujard, Bengali, Sunsilk, Queen Elizabeth, Julia’s Rose, Dutch Gold, King’s Ransom \cite{b2}. Not only from outer look, these rose breeds are different from inner part and that’s why they need different care for cultivation. Breed detection will be very helpful in the sector of rose cultivation and flowers business. As roses are the most common and demanded flower so people who cultivate flowers give more attention on cultivation of roses. According to the demand and the amount of rose cultivation, breeds detection of roses will be very beneficial. By knowing the breeds, it will be easier to detect the cultivating process and also the proper way of taking care of rose plants and flowers. All the rose breeds which are presently available in Bangladesh are basically hybrid breeds, so by detecting the rose breeds and comparing the pros and cons of all the breeds it will be possible for flower analysts to create or invent new rose breeds according to the climate of Bangladesh. Rose breed detection will play a vital role in rose cultivation. If cultivators will be aware about rose breeds, they will be educated about the cultivation of each breed. They will act based on the characteristics of the breeds, like the amount of fertilizer they need, how many days they take to grow, the limitation of a breed, caring method and many more. The two main goals of this research are as follows:
\begin{itemize}
    \item Develop an expert breed detection framework that can process the images of rose breeds and recognize the category of a particular breed.
    \item Maximized the data set amount by applying some image augmentation techniques, background noise removal, and data simulation.
\end{itemize}
The rest of the portion of this research paper is organized as follows. Section II summarized the important literature reviews. Proposed methodology along with data set distribution are described in section III. Experimental result evaluated and discussed in the Section IV. Finally we have concluded our work and narrated the future scope in section V.
\section{Related Works}
As per study on previous work, there is no work that has been presented on breed detection of a particular flower. There is various work on flower classification and identification of flower species. Some literature studies are below:\\
\\Philipe A. Dias et al.\cite{b3} have proposed a CNN based apple flower detection system. They have collected a total 147 images apple trees. To perform feature extraction, they have used fully connected layer of CNN and finally classified the images using SVM. Their CNN+SVM model achieved more than 90\% in terms of recall and precision score.\\
\\Xiaoling Xia et al.\cite{b4} classified flowers using the transfer learning technique based on Inception-v3 model. To classify flowers, they have used one data-set contains 17 species of flowers and for each species they have collected 80 flower images and another one contains 102 species of flowers and they have collected 40-258 flower images for each species. Their model obtained the classification accuracy 95\% on first data-set and on second data-set they have got 94\%.\\
\\I.Gogul et al.\cite{b5} proposed a system to recognize flower species using CNN. They have collected two datasets. For first dataset, they have collected 2240 images of flowers from 28 categories and for second dataset, they have collected 8189 images of flowers from 102 categories. They have used transfer learning approach for feature extraction. The highest accuracy of their model was 93.41\%.\\
\\Hazem Hiary et al. \cite{b6} represented two step deep learning classifier to identify flowers. They have modeled a binary classifier in a fully convolutional neural network framework and sturdy CNN classifier to identify the different flower types. They have used three datasets. The datasets have 8189 images from 102 categories, 1360 images from 17 categories and 612 images from 102 categories flowers respectively. They have got the classification outcome more than 97\% on every dataset.\\
\\M. Cıbuk et al. \cite{b7} used a hybrid method based on DCNN. For features extraction, they have combined the features from AlexNet and VGG16 models. At the end, they have applied mRMR method to specify more effective features. To identify the flower species by using the extracted features, they have applied the SVM classifier with RBF kernel. They have used two datasets of 17 and 102 flower categories accordingly. They obtained the accuracy rate 96.39\% and 95.70\% for the datasets individually. \\
\\Thi Thanh Nhan Nguyen et al.\cite{b8} tshowed the sturdiness of DCNN to identify flower species. They have extracted a flower dataset of 967 species from PlantCLEF 2015. The accuracy exceeded 90\% for that dataset.\\
\\Yuanyuan Liu et al.\cite{b9} prepared a method based on CNN for classification of flowers. They have created a dataset of 52,775 images from 79 species of flowers. By using this dataset, they have got 76.54\% of accuracy and they applied their method on a well-known dataset and got 84.02\% of accuracy.\\
\\Saiful Islam et al.\cite{b10} have developed a system to classify local flowers using CNN. Their dataset has 6400 images which belongs to eight kind of flower species. Their system has scored 85\% classification accuracy.\\
\\Busra Rumeysa Mete et al.\cite{b11} have classified flowers by using Deep CNN and machine learning approaches. They have used Deep CNN to perform feature extraction and to get better outcome they have showed the utilization of image augmentation. Finally, they have compared some machine learning approaches. They have used two datasets. They have achieved the highest accuracy 98.5\% and 99.8\% by using SVM and MLP classifier respectively.\\
\\Yong Wu et al.\cite{b12} have built a method to recognize flowers using CNN and transfer learning technic. For feature extraction, they applied DCNN. They have combined neural network model with transfer learning technic. By achieving a better accuracy of flower classification than other traditional methods, they have showed a remarkable enhancement.\\
\\S. M. Farhan Al Haque et al.\cite{b13} proposed a system for the detection of guava disease using deep learning models. Their final model achieved 95.61\% of accuracy.\\
\\The analysis of previous research works found that various researchers have applied different techniques to detect different fruits and vegetables. But nobody worked on flower classification and identification of flower species. We aim to predict rose breeds that can help the farmers and agriculturists to maximize the production of different roses.
\section{Proposed Methodology}
\begin{figure}[htbp]
\centerline{\includegraphics[width=0.45\textwidth]{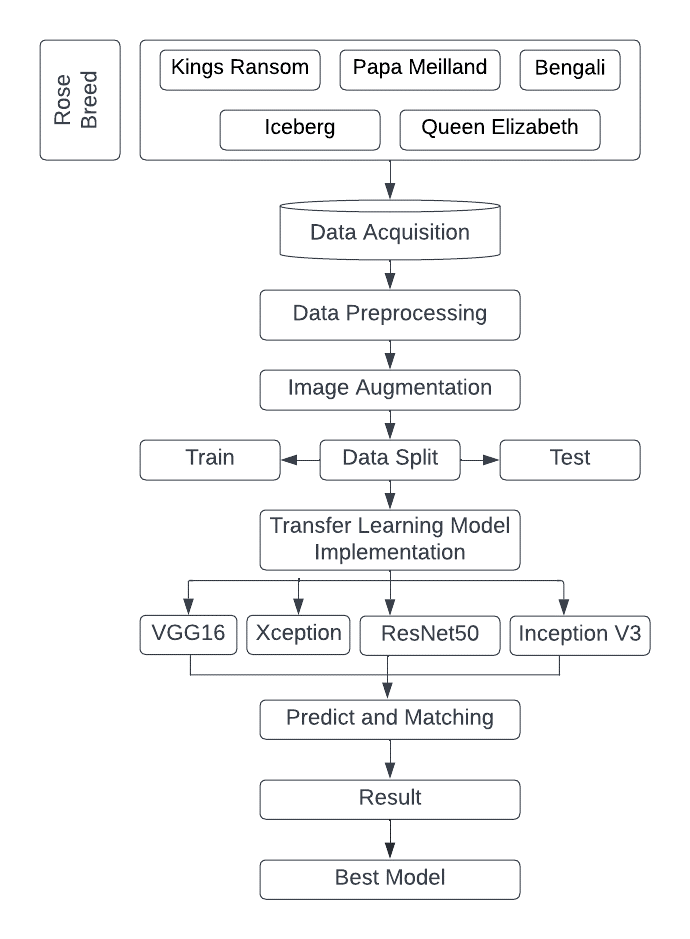}}
\caption{Working Diagram of Proposed Methodology}
\label{fig1}
\end{figure}

In this portion, we have found the model, which performed better for the dataset that we used to detect selected rose breeds. The presented working methodology has shown in  ``Fig.~\ref{fig1}''.

\subsection{Acquisition of Images }
\begin{figure}[htbp]
\centerline{\includegraphics[width=0.45\textwidth]{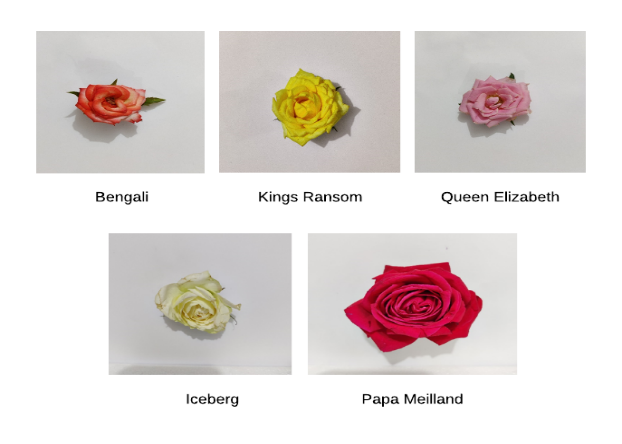}}
\caption{Working Diagram of Proposed Methodology}
\label{fig2}
\end{figure}

For a researched based work like this, data is the main and the central aspect to achieve a better outcome. So, we collected every single data manually and carefully for this research. We captured 1939 images of five different breeds of rose shown in``Fig.~\ref{fig2}''. These are Papa Meilland, Queen Elizabeth, Iceberg, Kings Ransom and Bengali. And we gathered 328, 465, 365, 391 and 390 images of these five breeds accordingly.\\
\subsection{Data Preprocessing and Data Augmentation}
Divided the dataset into two parts. 80\% of total collected data as training dataset and rest of the 20\% of the total data is for testing dataset. We used image augmentation in our training dataset to minimize overfitting problem. Also, it has been used to obtain higher accuracy in prediction and for overfitting of data. In this research rescale, horizontal flip, vertical flip, shear and zooming  augmentation methods had been used for the training dataset \cite{b14}. 
\begin{figure}[htbp]
\centerline{\includegraphics[width=0.45\textwidth]{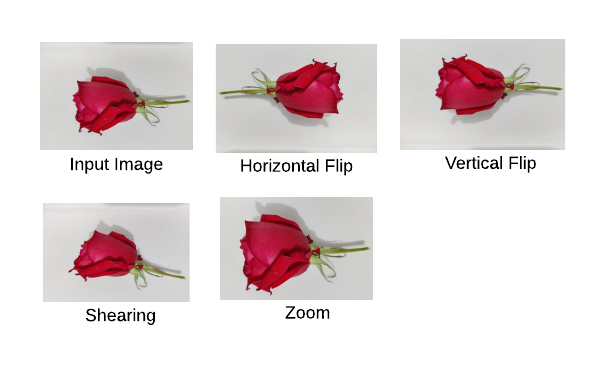}}
\caption{Images After Augmentation}
\label{fig3}
\end{figure}
We applied five image generation techniques \cite{b15} for training data processing and one image generation technique for testing data. Distribution of training dataset shown in ``Table.~\ref{table1}''
\begin{table}[htbp]
 \caption{Distribution of Training Dataset}
\label{table1}
\centering
\begin{tabular}{|c|c|c|c|}
\hline
Name of Class      & \begin{tabular}[c]{@{}l@{}}Collected\\Data(cd)\end{tabular} & \begin{tabular}[c]{@{}l@{}}Generated\\Data(gd)\end{tabular} & \begin{tabular}[c]{@{}l@{}}Total Data\\ (cd + gd) \end{tabular}\\ \hline
Papa Meilland      & 262    & 1310    & 1572    \\ \hline
Iceberg            & 292    & 1460    & 1752    \\ \hline
Kings Ransom       & 313    & 1565   & 1878    \\ \hline
Queen Elizabeth    & 372    & 1860    & 2232    \\ \hline
Bengali            & 312    & 1560    & 1872     \\ \hline
\end{tabular}
\end{table}

\subsection{Model Implementation }
For the dataset that we have collected, we tried to train an accurate model. Inception V3, Xception, ResNet 50 and VGG16 transfer learning models \cite{b16}, that we have used in this study. The basic theory of these models are described as follow:

Inception V3: Inception V3 \cite{b17} is one of the prominent deep learning models based on the Convolutional Neural Network (CNN). Inception V1 and V2 are the previous models of the same category of inception V3. Inception V3 is the latest version, which has higher efficiency, is comparatively less expensive in terms of computation, and uses auxiliary Classifiers as regularizes. In addition, it has 42 layers and a lower error rate.

Xception: Xception \cite{b18} means extreme inception. It is actually better than inception v3. The reverse process of the inception is known as xception. In the first step, filters are used on each depth map and in the second step it compresses the input space by applying it throughout the depth using 1x1 convolution. Inception and Xception are different from each other. The presence or absence of a non-linearity after the first operation. In the Inception model, both operations are followed by a ReLU non-linearity. On the other hand, Xception does not instigate any non-linearity.

ResNet50: ResNet50 \cite{b19} is a sub-model or a transformation of the ResNet model which includes 48-50 Convolution layers along with 1 MaxPool layer and 1 Average Pool layer. It makes the training time shorter than other variation of ResNet like ResNet 34 and ResNet 101, which makes it different. Instead of 2 it uses a stack of 3 layers. It assures more precise results in shorter time.

VGG16: VGG16 \cite{b20} is an intelligible model which is vastly used deep learning model used for ImageNet, a massive database project used in visual object recognition software research, arranged following to the WorldNet hierarchy. This model gains 92.7\% top5 accuracy in terms of testing in ImageNet, where a dataset consists of more than 14 million images that has 1000 classes. So, it can be said that, for visual recognition, it is an excellent architecture. VGG16 contains 16 layers, the size of RGB image from input to convolution layer 1 is fixed, which is 224 x 224. VGG16 focused on having convolution layers of a 3x3 filter with a stride one and always used the same padding and max pool layer of 2x2 filter of stride 2, instead of having a huge number of hyper-parameters, which is the most advantageous thing about VGG16. It usually follows this approach of convolution and max pool layers throughout the system. At last, it contains two fully connected layers followed by a SoftMax for output. This model has some drawbacks, it is quite time consuming to train the model and along around 138 million parameters, it is a giant network.

\section{Result and Discussion}
\begin{figure}[htbp]
\centerline{\includegraphics[width=0.4\textwidth]{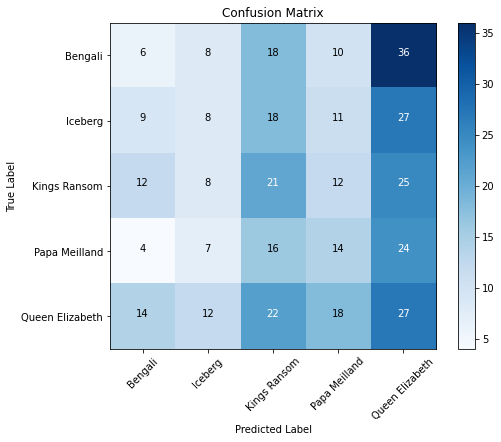}}
\caption{Confusion matrix for Inception V3}
\label{fig4}
\end{figure}

\begin{figure}[htbp]
\centerline{\includegraphics[width=0.4\textwidth]{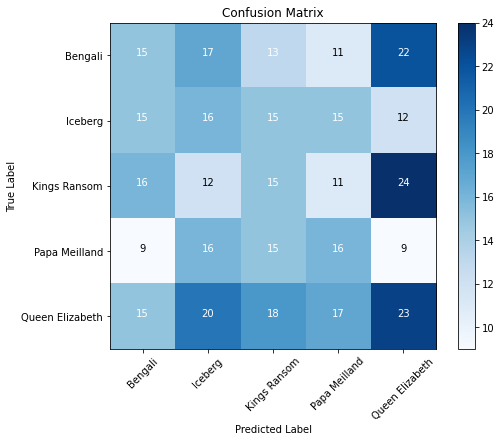}}
\caption{Confusion matrix for Xception}
\label{fig5}
\end{figure}

\begin{figure}[htbp]
\centerline{\includegraphics[width=0.4\textwidth]{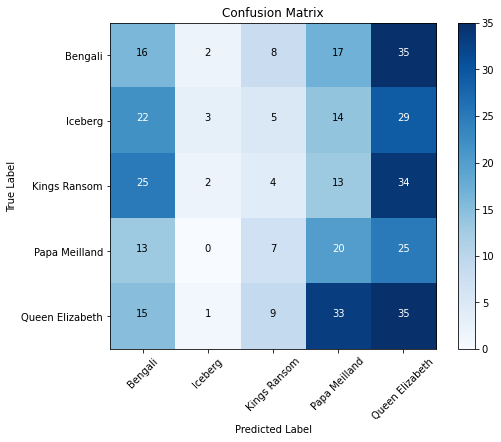}}
\caption{Confusion matrix for Resnet 50}
\label{fig6}
\end{figure}

\begin{figure}[htbp]
\centerline{\includegraphics[width=0.4\textwidth]{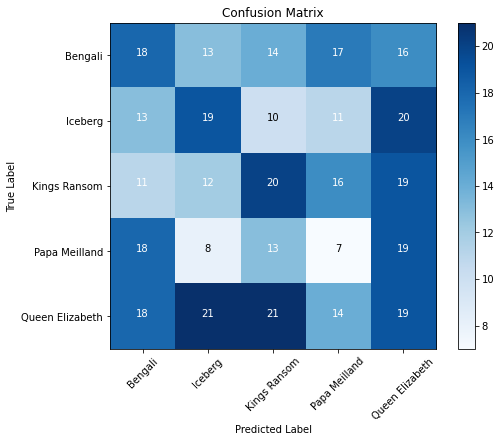}}
\caption{Confusion matrix for VGG16}
\label{fig7}
\end{figure}

\begin{figure}[htbp]
\centerline{\includegraphics[width=0.4\textwidth]{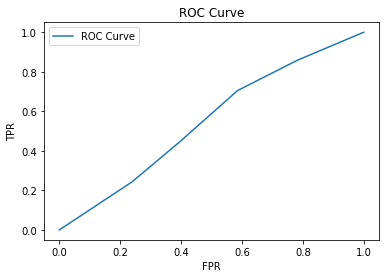}}
\caption{ROC curve for VGG 16}
\label{fig8}
\end{figure}

In this research, to train the models, we used 1550 images of five classes and to test the model, we used 387 images. We compared the confusion matrix of different algorithms to evaluate the accuracy. In this study, we have used four models. These are Inception V3, Xception, ResNet50 and VGG16. These are the most popular and used models in image identification and classification. To get a comprehensive knowledge, we have researched and compared the outcome of each transfer learning models. For each model, we have tried 20 epochs. To evaluate the performance of the models we have used confusion matrix. As there are a total five classes in our dataset, we used 5x5 matrix to calculate. ``Fig.~\ref{fig4}'', ``Fig.~\ref{fig5}'', ``Fig.~\ref{fig6}'', and ``Fig.~\ref{fig7}'' shows the confusion matrix for the Inception V3, Xception, Resnet 50 and VGG16 models respectively.
``Table.~\ref{table2}'' shows the accuracy and loss comparison for all the four models that we have applied. According to the ``Table.~\ref{table2}'' it is shows that VGG16 outperformed in terms of accuracy. This model shows the highest accuracy among the other three models. The ROC curve for the VGG16 model is shown in ``Fig.~\ref{fig8}'' . 
\begin{table}[htbp]
 \caption{Accuracy and Loss comparison of the Transfer Learning Models}
\label{table2}
\centering
\begin{tabular}{|c|c|c|c|c|}
\hline
\begin{tabular}[c]{@{}l@{}}Model\\Name\end{tabular}      & \begin{tabular}[c]{@{}l@{}}Training\\Accuracy\end{tabular} & \begin{tabular}[c]{@{}l@{}}Training\\Loss\end{tabular} & \begin{tabular}[c]{@{}l@{}}Test\\Accuracy \end{tabular} & \begin{tabular}[c]{@{}l@{}}Test\\Loss\end{tabular}\\ \hline

Inception V3    & 98.90\%    & 3.28\%    & 82.17\%  & 82.14\%   \\ \hline
VGG16            & 98.94\%    & 0.98\%    & 99\%     & 0.55\%  \\ \hline
ResNet50         & 68.71\%    & 82.97\%   & 70.28\%   & 68.08\% \\ \hline
Xception         & 99.68\%    & 1.63\%    & 96.64\%   & 17.28\% \\ \hline
\end{tabular}
\end{table}

\begin{figure}[htbp]
\centerline{\includegraphics[width=0.4\textwidth]{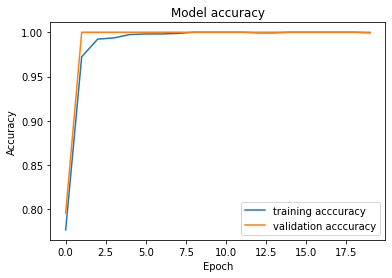}}
\caption{Model Accuracy for VGG16}
\label{fig9}
\end{figure}

\begin{figure}[htbp]
\centerline{\includegraphics[width=0.4\textwidth]{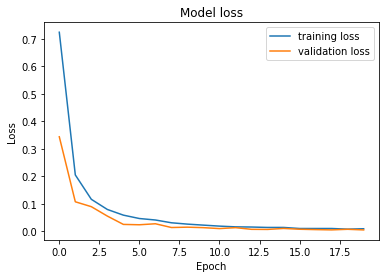}}
\caption{Model Loss for VGG16}
\label{fig10}
\end{figure}

The accuracy of model VGG16 and the model loss are shown in ``Fig.~\ref{fig9}'' and ``Fig.~\ref{fig10}'' respectively. In ``Table.~\ref{table3}'', we have shown a comparative analysis of our work with some other researches regarding  flower breed detection.

\begin{table*}[htbp]
\caption{Comparison between our work and recent researches}
\label{table3}
\centering
\begin{tabular}{|l|l|l|l|l|}
\hline
~~ List of Works & \begin{tabular}[c]{@{}l@{}}Working Topic\end{tabular} & \begin{tabular}[c]{@{}l@{}}Proposed Method\end{tabular} & \begin{tabular}[c]{@{}l@{}}Dataset Size\end{tabular} &\begin{tabular}[c]{@{}l@{}} Accuracy(\%)\end{tabular}\\ \hline

This work & \begin{tabular}[c]{@{}l@{}}Rose Breed detection\end{tabular} & \begin{tabular}[c]{@{}l@{}} Transfer Learning Technique\end{tabular} &  \begin{tabular}[c]{@{}l@{}} 1939 \end{tabular} &  \begin{tabular}[c]{@{}l@{}}99 \end{tabular} \\ \hline

Philipe A. Dias et al.\cite{b3} & \begin{tabular}[c]{@{}l@{}}Apple flower detection \end{tabular} & \begin{tabular}[c]{@{}l@{}} CNN + SVM \end{tabular} &  \begin{tabular}[c]{@{}l@{}} 147 \end{tabular} &  \begin{tabular}[c]{@{}l@{}}90 \end{tabular} \\ \hline

I.Gogul et al.\cite{b5} & \begin{tabular}[c]{@{}l@{}}Flower species recognition\end{tabular} & \begin{tabular}[c]{@{}l@{}} CNN\end{tabular} &  \begin{tabular}[c]{@{}l@{}} 8189 \end{tabular} &  \begin{tabular}[c]{@{}l@{}}93.41 \end{tabular} \\ \hline

Yuanyuan Liu et al.\cite{b9} & \begin{tabular}[c]{@{}l@{}}Flower classification\end{tabular} & \begin{tabular}[c]{@{}l@{}} CNN\end{tabular} &  \begin{tabular}[c]{@{}l@{}} 52775 \end{tabular} &  \begin{tabular}[c]{@{}l@{}}76.54 \end{tabular} \\ \hline

Saiful Islam et al. \cite{b10} & \begin{tabular}[c]{@{}l@{}}Local flowers classification\end{tabular} & \begin{tabular}[c]{@{}l@{}} CNN\end{tabular} &  \begin{tabular}[c]{@{}l@{}} 6400 \end{tabular} &  \begin{tabular}[c]{@{}l@{}}85 \end{tabular} \\ \hline

\end{tabular}
\end{table*}

\section{Conclusion And Future Work}
\begin{figure}[htbp]
\centerline{\includegraphics[width=0.4\textwidth]{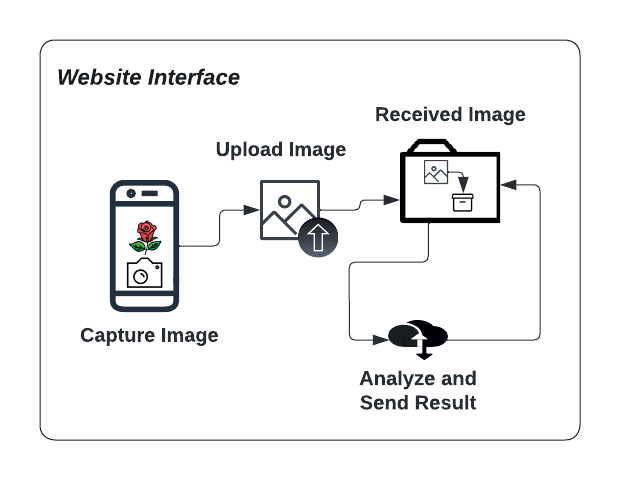}}
\caption{Proposed web-based architecture of rose breed detection}
\label{fig11}
\end{figure}

In this research work, we have proposed and discussed about a significant Algorithm for breed detection. In addition, we have used a real-world dataset which contains 19,39 images of five different breeds of rose. we have applied four Transfer Learning methods, these are Inception V3, VGG16, ResNet50 and Xception. Among these four methods, the performance of VGG16 was great and achieved the highest accuracy of 99\%. 
We would like to build a Transfer Learning methods-based system in future. This system will not only detect the breed of a rose but also it will provide the characteristics of that particular breed and the cultivation and caring process of the breed. ``Fig.~\ref{fig11}'' shows the proposed web-based architecture of rose breed detection.

In this system, at first a user needs to capture a picture of a rose or upload a picture. Then, when the picture will be sent to the system, the system will work on it. It will analysis the picture and it will compare the breed with the breeds which are present in the database. If it gets any similarity in between the input data and the data of the database, it will display an output with the name of the breed of the rose with the details of that breed and also will provide the proper way of cultivating and taking care of the breed. As we have used transfer learning techniques instead of traditional CNN, the system is complex and more time consuming.

\vspace{12pt}


\begin{thebibliography}{00}
\bibitem{b1} Mou, N. H. (2012). Profitability of Flower Production and Marketing System of Bangladesh. Bangladesh Journal of Agricultural Research, 37(1), 77–95. https://doi.org/10.3329/bjar.v37i1.11179
\bibitem{b2} Kabir, Ashraful and Rahman, Ataur. (2020). "Rose Varieties and Their Caring in Bangladesh."
\bibitem{b3} P. A. Dias, A. Tabb, and H. Medeiros, “Apple flower detection using deep convolutional networks,” Computers in Industry, vol. 99, pp. 17–28, 2018.
\bibitem{b4} Xiaoling Xia, Cui Xu and Bing Nan, "Inception-v3 for flower classification,"2017 2nd International Conference on Image, Vision and Computing (ICIVC), pp. 783-787, 2017.
\bibitem{b5} I. Gogul and V. S. Kumar, "Flower species recognition system using convolution neural networks and transfer learning," 2017 Fourth International Conference on Signal Processing, Communication and Networking (ICSCN), pp. 1-6, 2017. 
\bibitem{b6} Hiary, H., Saadeh, H., Saadeh, M. and Yaqub, M., “Flower classification using deep convolutional neural networks,” IET Computer Vision, 12(6), pp.855-862, 2018.
\bibitem{b7} M. Cıbuk, U. Budak, Y. Guo, M. Cevdet Ince and A. Sengur, "Efficient deep features selections and classification for flower species recognition,” Measurement, vol. 137, pp. 7-13, 2019.
\bibitem{b8} Thi Thanh Nhan Nguyen, Van Tuan Le, Thi Lan Le, Hai Vu, Natapon Pantuwong, and Yasushi Yagi, “Flower species identification using deep convolutional neural networks,” AUN/SEED-Net Regional Conference for Computer and Information Engineering 2016.
\bibitem{b9} Y. Liu, F. Tang, D. Zhou, Y. Meng and W. Dong, "Flower classification via convolutional neural network", 2016 IEEE International Conference on Functional-Structural Plant Growth Modeling, Simulation, Visualization and Applications (FSPMA), pp. 110-116, 2016.
\bibitem{b10} S. Islam, M. F. Ahmed Foysal and N. Jahan, "A Computer Vision Approach to Classify Local Flower using Convolutional Neural Network," 2020 4th International Conference on Intelligent Computing and Control Systems (ICICCS), pp. 1200-1204, 2020.
\bibitem{b11} B.R. Mete and T. Ensari, "Flower Classification with Deep CNN and Machine Learning Algorithms," 2019 3rd International Symposium on Multidisciplinary Studies and Innovative Technologies (ISMSIT), pp. 1-5, 2019. 
\bibitem{b12} Y. Wu, X. Qin, Y. Pan and C. Yuan, "Convolution Neural Network based Transfer Learning for Classification of Flowers," 2018 IEEE 3rd International Conference on Signal and Image Processing (ICSIP), pp. 562-566, 2018. 
\bibitem{b13} A. S. M. Farhan Al Haque, R. Hafiz, M. A. Hakim and G. M. Rasiqul Islam, "A Computer Vision System for Guava Disease Detection and Recommend Curative Solution Using Deep Learning Approach," 2019 22nd International Conference on Computer and Information Technology (ICCIT), 2019, pp. 1-6, doi: 10.1109/ICCIT48885.2019.9038598.
\bibitem{b14} A. S. M. F. Al Haque, M. A. Hakim and R. Hafiz, "CNN Based Automatic Computer Vision System for Strain Detection and Quality Identification of Banana," 2021 International Conference on Automation, Control and Mechatronics for Industry 4.0 (ACMI), 2021, pp. 1-6, doi: 10.1109/ACMI53878.2021.9528269. 
\bibitem{b15} S. D. Ray, M. K. T. K. Natasha, M. A. Hakim and F. Nur, "Carrot Cure: A CNN based Application to Detect Carrot Disease," 2022 6th International Conference on Trends in Electronics and Informatics (ICOEI), 2022, pp. 01-07, doi: 10.1109/ICOEI53556.2022.9776947.
\bibitem{b16} S. Hasan, G. Rabbi, R. Islam, H. Imam Bijoy and A. Hakim, "Bangla Font Recognition using Transfer Learning Method," 2022 International Conference on Inventive Computation Technologies (ICICT), 2022, pp. 57-62, doi: 10.1109/ICICT54344.2022.9850765.
\bibitem{b17} Xiaoling Xia, Cui Xu and Bing Nan, "Inception-v3 for flower classification," 2017 2nd International Conference on Image, Vision and Computing (ICIVC), 2017, pp. 783-787, doi: 10.1109/ICIVC.2017.7984661.
\bibitem{b18} W. W. Lo, X. Yang and Y. Wang, "An Xception Convolutional Neural Network for Malware Classification with Transfer Learning," 2019 10th IFIP International Conference on New Technologies, Mobility and Security (NTMS), 2019, pp. 1-5, doi: 10.1109/NTMS.2019.8763852.
\bibitem{b19} Y. Wu, X. Qin, Y. Pan and C. Yuan, "Convolution Neural Network based Transfer Learning for Classification of Flowers," 2018 IEEE 3rd International Conference on Signal and Image Processing (ICSIP), 2018, pp. 562-566, doi: 10.1109/SIPROCESS.2018.8600536.
\bibitem{b20} D. I. Swasono, H. Tjandrasa and C. Fathicah, "Classification of Tobacco Leaf Pests Using VGG16 Transfer Learning," 2019 12th International Conference on Information \& Communication Technology and System (ICTS), 2019, pp. 176-181, doi: 10.1109/ICTS.2019.8850946.

\end{thebibliography}
\end{document}